\pdfoutput=1
\documentclass[11pt]{article}

\usepackage{ACL2023}

\usepackage{times}
\usepackage{latexsym}

\usepackage[T1]{fontenc}
\usepackage[utf8]{inputenc}

\usepackage{microtype}

\usepackage{inconsolata}

\usepackage{graphicx}
\usepackage{enumitem}
\usepackage{caption, subcaption}
\usepackage{fancyvrb}
\usepackage{amsmath}
\usepackage{amssymb}
\usepackage{multirow}
\usepackage{booktabs}
\usepackage{algorithm}
\usepackage{algpseudocode}  
\usepackage{color}

\title{Universal Information Extraction with Meta-Pretrained Self-Retrieval}

\author{Xin Cong$^{1,2}$ Bowen Yu$^{1,2}$ \ Mengcheng Fang$^{3}$ \ \bf{Tingwen Liu}$^{1,2*}$ \ Haiyang Yu$^{3}$ \\ \bf{Zhongkai Hu}$^{3}$ \ \bf{Fei Huang}$^{3}$ \ \bf{Yongbin Li}$^{3}$\thanks{\hspace{0.15cm}Corresponding   Author} \and \bf{Bin Wang}$^{4}$  \\
	$^1$Institute of Information Engineering, Chinese Academy of Sciences. Beijing, China \\
	$^2$School of Cyber Security, University of Chinese Academy of Sciences. Beijing, China \\
	$^3$DAMO Academy, Alibaba Group \\
	$^4$Xiaomi AI Lab, Xiaomi Inc., Beijing, China \\
	{\tt \{congxin, liutingwen\}@iie.ac.cn} \\
	{\tt \{yubowen.ybw, yifei.yhy, zhongkai.hzk, f.huang, shuide.lyb\}@alibaba-inc.com} \\
	{\tt fmc1121@zju.edu.cn} \\ 
	{\tt wangbin11@xiaomi.com}
}

\begin{document}
	\maketitle
	\begin{abstract}
		Universal Information Extraction~(Universal IE) aims to solve different extraction tasks in a uniform text-to-structure generation manner.
		Such a generation procedure tends to struggle when there exist complex information structures to be extracted.
		Retrieving knowledge from external knowledge bases may help models to overcome this problem but it is impossible to construct a knowledge base suitable for various IE tasks. 
		Inspired by the fact that large amount of knowledge are stored in the pretrained language models~(PLM) and can be retrieved explicitly, in this paper, we propose MetaRetriever to retrieve task-specific knowledge from PLMs to enhance universal IE.
		As different IE tasks need different knowledge, we further propose a Meta-Pretraining Algorithm which allows MetaRetriever to quicktly achieve maximum task-specific retrieval performance when fine-tuning on downstream IE tasks.
		Experimental results show that MetaRetriever achieves the new state-of-the-art on 4 IE tasks, 12 datasets under fully-supervised, low-resource and few-shot scenarios.
	\end{abstract}
	
	\section{Introduction}
	
	Information extraction (IE) is the task of extracting specific information structures of interest from unstructured text~\citep{andersen-etal-1992-automatic,grishman_2019}. 
	The various IE tasks at hand can be highly heterogeneous, as each task may have different extraction targets, such as entities~\citep{lample-etal-2016-neural}, relations~\citep{zheng-etal-2017-joint}, and events~\citep{lin-etal-2018-nugget}. Traditional IE methods tend to be task-specific, resulting in specialized and isolated approaches for different tasks.
	
	In recent years, researchers have made efforts towards universal information extraction, or Universal IE~\citep{uie2022lu}, which aims to develop a unified method capable of solving a range of IE tasks. 
	One proposed approach involves the use of a Structure Extraction Language to express different extraction targets in a unified structure form, and pretrains a Transformer model to generate this unified structure through a sequence-to-sequence process. 
	However, this structure generation process can be limited in its ability to fully utilize contextual semantic correlation between the extracted information, as the Transformer generation process is unidirectional. 
	In cases where the information structure to be extracted is complex, the universal IE model may struggle to generate accurate results.

	Previous studies on retrieval-augmented generation~\citep{lewis2020retrieval,DBLP:conf/acl/WangJBWHHT20,cai2022recent,DBLP:journals/corr/abs-2202-09022} have shown that utilizing external knowledge bases to retrieve task-specific information can improve the ability of models to generate complex sequences. 
	However, this approach is not practical for Universal IE, as it is almost impossible to build a knowledge base suitable for various IE tasks, and it is also the ultimate goal of IE.
	Recently, research on knowledge probing~\citep{DBLP:conf/emnlp/PetroniRRLBWM19,DBLP:conf/emnlp/RobertsRS20,zhang2022automatic,yu2022generate} has demonstrated that large amounts of knowledge are stored in pretrained language models (PLMs) and can be retrieved explicitly. 
	Utilizing this knowledge to enhance IE models could avoid the need for extensive resource construction. 
	In light of these findings, the question arises: \textit{can PLMs be used as knowledge bases to retrieve knowledge and improve universal IE models?} If so, universal IE models would be able to generate more accurate results.

	Based on this idea, in this paper, we propose MetaRetriever, a pretrained model for universal IE.
	Unlike existing method which generates results in a single step, MetaRetriever utilizes a retrieve-then-extract approach: first, it retrieves task-specific knowledge from itself, and then it uses this knowledge as additional input to extract information.
	However, different IE tasks require different knowledge, and simply retrieving information from PLM can lead to irrelevant information being included, which can negatively impact performance.
	To address this, incorporating meta-learning techniques~\citep{DBLP:conf/icml/FinnAL17}, we develop a Meta-Pretraining Algorithm~(MPA) which allows the model to quickly learn the semantics of extraction targets of different tasks to retrieve the relevant knowledge.
	Specifically, MPA pretrains the model with a bi-level optimization, in which the model are optimized to achieve maximal task-specific retrieval performance given the extraction targets after the parameters have been updated through a small number of gradient steps on a new IE task.
	Thus, in the fine-tuning phase, MetaRetriever can learn downstream tasks quickly to retrieve relevant knowledge and disregard irrelevant information.
	
	\begin{figure}[t] 
		\centering
		
		\begin{subfigure}[b]{0.45\textwidth}
			\begin{minipage}{1.0\textwidth}
				\centering
				\includegraphics[width=1.0\linewidth]{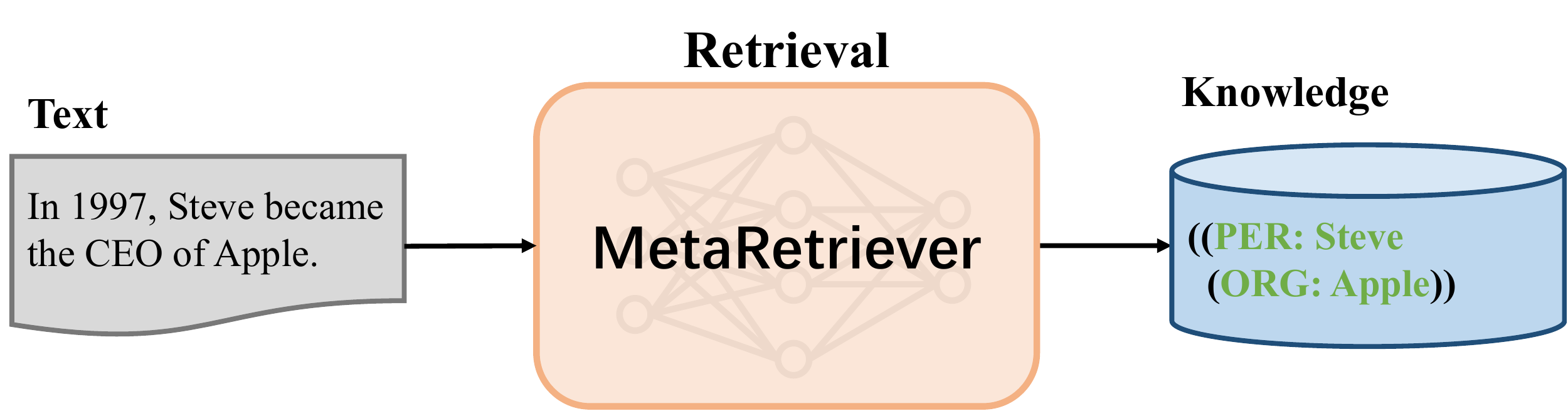}
			\end{minipage}
			\caption{
				First: Retrieval task-specific knowledge from the model.
			}
			\label{fig:model_retrieval}
		\end{subfigure}
		
		\begin{subfigure}[b]{0.45\textwidth}
			\begin{minipage}{1.0\textwidth}
				\centering
				\includegraphics[width=1.0\linewidth]{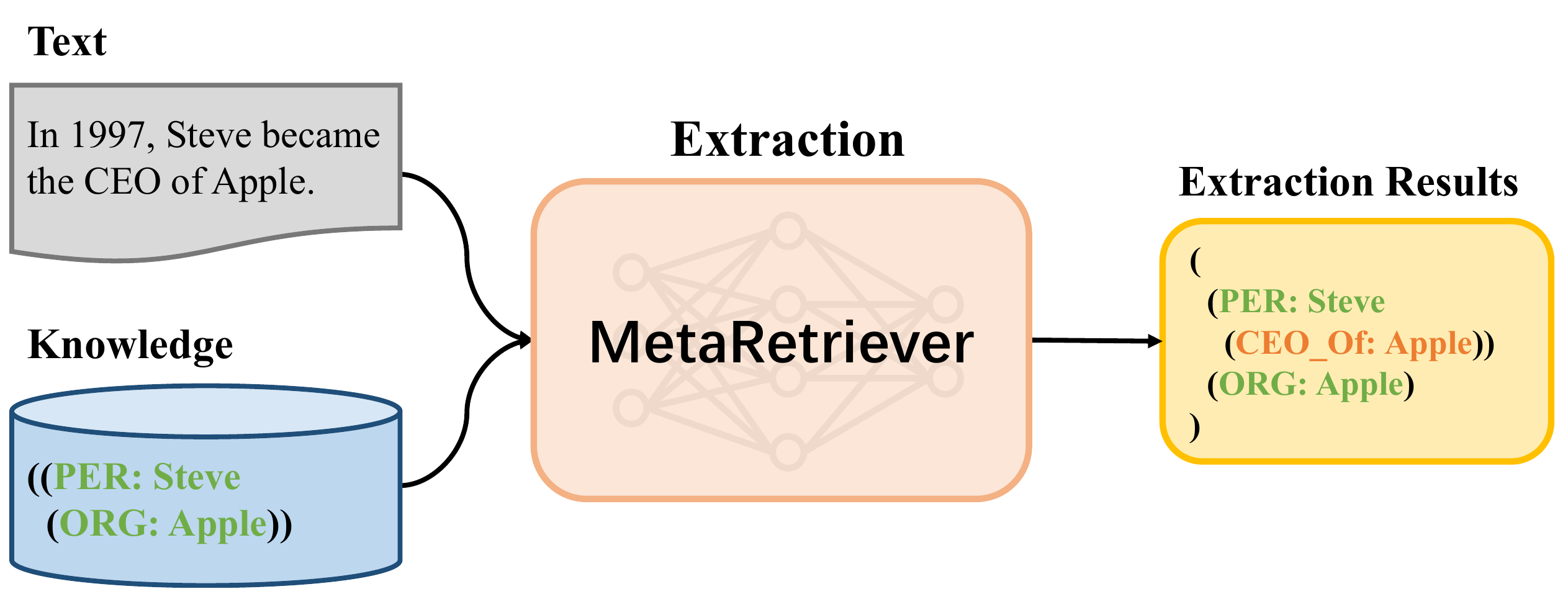}
			\end{minipage}
			\caption{
				Second: Extraction based on the retrieved knowledge.
			}
			\label{fig:model_generation}
		\end{subfigure}
		
		\setlength{\belowcaptionskip}{-0.4cm}
		\caption{
			Illustrations of our proposed MetaRetriever with the retrieve-then-extract manner.
		}
		
		\label{fig:metaretriever}
	\end{figure}

	We conduct experiments on 12 datasets of 4 main IE tasks under fully-supervised, few-shot and low-resource scenarios.
	Experiments show that MetaRetriever significantly surpasses previous works and achieves the new state-of-the-art.
	On fully-supervised scenario, MetaRetriever achieves 0.7\% gains on average over UIE~\citep{uie2022lu}.
	On few-shot and low-resource scenarios, our model outperforms UIE with over 2.0\% improvement.
	
	Our contributions can be summarized as follows:
	(1) We introduce a retrieve-then-extract perspective into universal IE. It retrieves task-specific knowledge from the pretrained language model to improve performance.
	(2) We propose a meta-pretraining algorithm to make models fast-adapt into various downstream IE tasks by retrieving task-specific knowledge.
	(3) We construct large-scale pretraining corpus for universal IE and it can promote future research\footnote{Our code and constructed corpus are released at \url{https://github.com/AlibabaResearch/DAMO-ConvAI/tree/main/metaretriever}}.

	\section{Preliminaries}
	
	\begin{figure}[t]
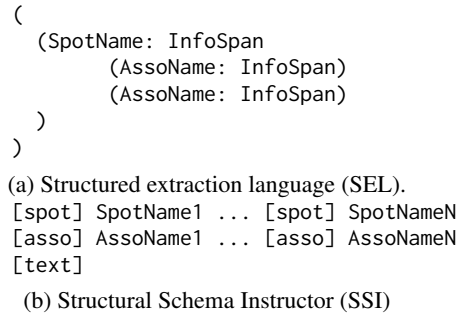
 
		\centering
		
		\begin{subfigure}[b]{0.48\textwidth}
			\begin{minipage}{.48\textwidth}
				\centering
				\begin{Verbatim}[fontsize=\small]
	(
	  (SpotName: InfoSpan
		(AssoName: InfoSpan)
		(AssoName: InfoSpan)
	  )
	)
				\end{Verbatim}
			\end{minipage}
			\caption{
				Structured extraction language (SEL).
			}
			\label{fig:sel}
		\end{subfigure}
		
		%	\vfill
		
		\begin{subfigure}[b]{0.48\textwidth}
			\begin{minipage}{.96\textwidth}
				\centering
				\begin{Verbatim}[fontsize=\small,commandchars=\\\{\}]
	[spot] SpotName1 ... [spot] SpotNameN 
	[asso] AssoName1 ... [asso] AssoNameN 
	[text] 
				\end{Verbatim}
			\end{minipage}
			\caption{
				Structural Schema Instructor (SSI)
			}
			\label{fig:ssi}
		\end{subfigure}
		
		\setlength{\belowcaptionskip}{-0.4cm}
		\caption{
			Illustrations of SEL and SSI.
		}
		
		\label{fig:ssi_sel}
	\end{figure}

	\subsection{Structured Extraction Language}
	
	The UIE paper~\citep{uie2022lu} presents the concept of the Structured Extraction Language (SEL) for expressing the diverse extraction targets of various IE tasks in a uniform representation. 
	As illustrated in \figurename~\ref{fig:sel}, SEL is a hierarchical key-value structure that comprises three components: 
	
	(1) \texttt{SpotName}: the targeted class name of the extracted information pieces in a specific IE task, e.g., ``PERSON'' in named entity recognition;
	
	(2) \texttt{AssoName}: the targeted class name of the relationship between different information, e.g., ``Works For'' in relation extraction;
	
	(3) \texttt{InfoSpan}: the text span corresponding to the specific SpotName or AssoName.
	
	Additionally, the colon (``:'') symbol denotes the mapping of the text span to its class name or relationship, and the hierarchical structure is indicated by the use of parentheses (``('' and ``)'').

	\subsection{Structural Schema Instructor}
	
	To make the model be aware of which extraction targets the IE task focus on, UIE further proposes Structural Schema Instructor~(SSI), a schema-based prompt.
	As shown in \figurename~\ref{fig:ssi}, SSI first uses several special symbols ([spot], [asso]) to denote SpotNames and AssoNames. 
	Then these SpotNames and AssoNames are concatenated before each text sequence with a ``[text]'' symbol as a prompt.
	UIE takes the prompted text as input to generate the linearized SEL to achieve information extraction.

	\section{Methodology}

	Our proposed MetaRetriever achieves universal IE via a retrieve-then-extract manner. 
	It first retrieves task-specific knowledge from itself according to the input text and then generates the extraction results based on the retrieved knowledge.
	To make MetaRetriever only retrieve task-specific knowledge and neglect other distracting knowledge, we further proposed a meta-pretraining algorithm based on meta-learning technique.
	In this section, we first introduce how MetaRetriever works via the retrieve-then-extract manner and then detail the meta-pretraining algorithm.
	
	\subsection{Retrieve-then-Extract}

	Given that providing task-specific knowledge can be beneficial for IE and that pretrained models have acquired a significant amount of knowledge through training on large corpora~\citep{DBLP:conf/emnlp/PetroniRRLBWM19,DBLP:conf/emnlp/RobertsRS20,zhang2022automatic,yu2022generate}, 
	our proposed system, MetaRetriever, utilizes this pretrained knowledge by first retrieving relevant information from the model itself. 
	This retrieved knowledge then serves as an additional input, along with the original input, to produce the final extraction results. 
	The design of MetaRetriever is depicted in \figurename~\ref{fig:metaretriever}.
	
	\subsubsection{Retrieval Procedure}
	
	We use SEL~\citep{uie2022lu} to express knowledge for knowledge retrieval and knowledge understanding.
	To specify which extraction targets MetaRetriever should retrieve, we adopt SSI as a prefix that controls which kinds of knowledge need to be retrieved.
	Thus, the input is in the form of:
	\begin{equation}
		\small
		\begin{aligned}
			s \oplus x & = [s_{1}, s_{2}, ..., s_{|s|}, x_{1}, x_{2}, ..., x_{|x|}] \\
		\end{aligned}
	\end{equation}
	where $s = [s_{1}, s_{2}, ..., s_{|s|}]$ is SSI and $x = [x_{1}, x_{2}, ..., x_{|x|}]$ is the original text.
	%For example, the prompt ``[spot] person [spot] company [asso] work for [text]'' indicates extracting records of the relation schema ``the person works for the company'' from the sentence.
	%
	%Given the prompt $s$, MetaRetriever can be aware of which task-specific knowledge should be retrieved. 
	%
	Defining a unified format of task-specific knowledge for various IE tasks is a challenging issue. 
	In this paper, we adopt a simple strategy: using the ground truth linearized SEL sequence of the corresponding input text as the knowledge we wish to retrieve.
	% Since MetaRetriever aims to retrieve task-related knowledge based on the input text, we set the ground truth linearized SEL sequence of the corresponding input text as the golden retrieval answer.
	%
	Taking SSI ($s$) and the text sequence ($x$) as input, MetaRetriever generates the linearized SEL ($k$) to retrieve task-specific knowledge.
	%%
	%As shown in \figurename~\ref{}, each hierarchical key-value structure expression contains three types of semantic parts:
	%\begin{itemize}
	%	\item \textsc{SpotName} indicates there exists an extracted information piece with the type of SpotName;
	%	\item \textsc{AssoName} indicates there exists an extracted information piece that is with the AssoName association to its upper-level Spot information piece;
	%	\item \textsc{InfoSpan} indicates the text span corresponding to the specific SpotName or AssoName.
	%\end{itemize}
	%%
	%Furthermore, ``:'' indicates the mapping from InfoSpan to its SpotName or AssoName, and the two structure indicators ``('' and ``)'' are used to form the hierarchical structure between the extracted information.
	%
	The retrieval procedure can represented as follows:
	\begin{equation}
		\small
		k = \text{MetaRetriever}(s \oplus x)
	\end{equation}
	where $k = [k_{1}, ..., k_{|k|}] $ is the retrieved knowledge represented by SEL.
	
	\subsubsection{Extraction Procedure}
	
	After retrieval of task-specific knowledge, MetaRetriever incorporates this information as an additional input to generate the extraction results. 
	To ensure that MetaRetriever utilizes the retrieved knowledge in generating the results, we concatenate it with the original input by appending it as a suffix:
	\begin{equation}
		\small
		\begin{aligned}
			s \oplus x \oplus k & = [s_{1}, ..., s_{|s|}, x_{1}, ..., x_{|x|}, k_{1}, ..., k_{|k|}]
		\end{aligned}
	\end{equation}
	Then, MetaRetriever will take $s \oplus x \oplus k$ as the new input to generate the final extraction results $y$:
	\begin{equation}
		\small
		y = \text{MetaRetriever}(s \oplus x \oplus k)
	\end{equation}
	
	%Since MetaRetriever works in a sequence-to-squence manner, we use the encoder-decoder-style architecture to achieve this procedure (e.g., T5~\citep{DBLP:journals/jmlr/RaffelSRLNMZLL20} and BART~\citep{DBLP:conf/acl/LewisLGGMLSZ20}).

	\subsection{Meta-Pretraining Algorithm}
	
	\begin{algorithm}[!t]
		\small
		\caption{Meta-Pretraining Algorithm}
		\label{alg:pretrain}
		\begin{algorithmic}[1]  %1表示每隔一行编号	
			\Require $\alpha$,$\;\beta$: inner/outer step size; $\mathcal{D}$: pretraining corpus 
			\State Initialize parameters $\theta$ of the model
			\State Construct support-query pair based on $\mathcal{D}$
			
			\State \texttt{// Outer Loop}
			\While{not converged}
			\State Sample the support set $(x^S, y^S)$ 
			\Statex\quad \quad and the query set $(x^Q, y^Q)$
			
			\State Compute record loss $\mathcal{L}_{\text{record}}(y; \theta)$
			
			\State Corrupt raw text: $x', x'' \leftarrow \mathtt{Corrupt}(x)$
			\State Compute LM loss $\mathcal{L}_{\text{LM}}(x', x''; \theta)$
			
			\State \texttt{// Inner Loop}
			\For{$j$ in number of inner-loop updates $J$}
			\State Compute retrieval loss $\mathcal{L}_{\text{retrv}}(x^S, y^S; \theta_j)$
			
			\State Compute extraction loss $\mathcal{L}_{\text{ext}}(x^S, k^S, y^S; \theta_j)$
			
			\State Compute inner loss $\mathcal{L}_{\text{inner}} = \mathcal{L}_{\text{retrv}} + \mathcal{L}_{\text{ext}}$
			
			\State Compute adapted parameters 
			\Statex\qquad \qquad $\theta_{j+1} \leftarrow \theta_j-\alpha \nabla_{\theta_{j}}\mathcal{L}_{\text{inner}}$
			\EndFor
			
			\State Compute retrieval loss $\mathcal{L}_{\text{retrv}}(x^Q, y^Q; \theta_J)$
			\State Compute extraction loss $\mathcal{L}_{\text{ext}}(x^Q, k^Q, y^Q; \theta_J)$
			\State Compute outer loss
			\Statex\quad \quad \quad $\mathcal{L}_{\text{outer}} = \mathcal{L}_{\text{retrv}} + \mathcal{L}_{\text{ext}} + \mathcal{L}_{\text{record}} + \mathcal{L}_{\text{LM}}$
			
			\State Update model parameters $\theta \leftarrow \theta-\beta \nabla_{\theta} \mathcal{L}_{\text{outer}}$
			\EndWhile
		\end{algorithmic}
	\end{algorithm}

	As the knowledge contained in the pretrained model may not always be relevant to the downstream IE tasks, we propose a Meta-Pretraining Algorithm~(MPA) that utilizes a meta-learning technique~\citep{DBLP:conf/icml/FinnAL17} to enable MetaRetriever to quickly learn the semantics of the extraction targets of downstream IE tasks, and thus, retrieve task-specific knowledge only. The algorithm is detailed in Algorithm~\ref{alg:pretrain}.
	
	% Since knowledge stored in the pretrained model is not always related to the downstream IE tasks, we propose a meta-pretraining algorithm based meta-learning technique~\citep{DBLP:conf/icml/FinnAL17} to empower MetaRetriever to learn the extraction target semantics of downstream IE tasks quickly for retrieving task-specific knowledge only.
	%
	%In this section, we first introduce the meta-pretraing dataset construction and then present the meta-pretraining algorithm.
	%
	% The algorithm is outlined in Algorithm~\ref{alg:pretrain}.
	
	%\subsubsection{Meta-Pretraining Algorithm}
	
	% To make MetaRetriever have the fast learning ability to lean various IE tasks, we meta-pretrain MetaRetriever with a bi-level optimization of the meta-learning technique~\citep{DBLP:conf/icml/FinnAL17}.
	%
	To make MetaRetriever retrieve task-specific knowledge, MPA meta-pretrain MetaRetriever with a bi-level optimization~\citep{DBLP:conf/icml/FinnAL17} to optimize MetaRetriever to achieve maximum task-specific retrieval performance after the parameters have been updated through a small number of gradient steps on a new IE task.
	In the bi-level optimization, models will experience two optimization trials: the inner loop and the outer loop.
	The inner loop aims to mimic the downstream fine-tuning procedure to compute a intermediate parameters of the model via SGD~\citep{saad1998online,DBLP:conf/icml/SutskeverMDH13}.
	The outer loop aims to mimic the downstream evaluation procedure to optimize the model parameters based on the intermediate parameters computed in the inner loop.
	The inner loop and the outer loop are nested and thus, it calculates high-order gradients to update parameters.
	Optimized by such a bi-level optimization, MetaRetriever would lean that after finetuning a small number of steps (the inner loop), MetaRetriever can achieve lower retrieval loss (the outer loop), a.k.a. learning a fast-adaptation ability for task-specific retrieval.

	To accomplish this, MetaRetriever should be pretrained on a set of simulated ``IE tasks'' $\mathcal{T}$ in the pretraining phase.
	Each ``IE task'' contains a support set $\mathcal{S}$ and a query set $\mathcal{Q}$.
	The support set is to mimic the training set used in the inner loop and the query set is to mimic the test set used in the outer loop.
	To construct various ``IE tasks'', a intuitive way is to follow the widely-used \textit{episodic training} technique in conventional meta-learning methods~\citep{DBLP:conf/nips/VinyalsBLKW16} to randomly sample instances by category.
	However, such a random-sampling operation requires random access to all data, which is infeasible when processing large-scale corpora, as such corpora cannot be stored in memory and must be accessed from external storage instead, resulting in intolerable long input/output (IO) time.
	To address this limitation, we design a support-query pairing method based on the graph maximum-weighted matching algorithm~\citep{DBLP:journals/csur/Galil86}, which significantly reduces IO time (see in Section~\ref{sec:exp_graph_match}). 
	% 为什么time-consuming？需要解释
	%
	% Thus, we use the graph maximum-weighted matching algorithm~\citep{DBLP:journals/csur/Galil86} to construct support-query instead.
	% 并且我们用实验表明了我们的策略具有更好的速度
	Details of our method are introduced in Appendix~\ref{sec:pairing}.

	\paragraph{Inner Loop}
	Since the inner loop simulates the downstream fine-tuning process, MetaRetriever computes the intermediate parameters by considering two inner losses: the retrieval loss $\mathcal{L}_{\text{retrv}}$ and the extraction loss $\mathcal{L}_{\text{ext}}$. Let MetaRetriever be represented by a parametrized function $f_\theta$ with parameters $\theta$. When learning a mimic IE task $\mathcal{T}_i = \{ \mathcal{S}_i, \mathcal{Q}_i \}$, the retrieval loss is computed based on the support set $\mathcal{S}_i = (x^{\mathcal{S}_i}, y^{\mathcal{S}_i})$.
	\begin{equation}
		\small
		\mathcal{L}_\text{retrv} = \verb|CrossEntropy| \left(f_\theta(x^{\mathcal{S}_i}), y^{\mathcal{S}_i} \right)
	\end{equation}
	It takes the text with SSI as the input and calculates the cross-entropy loss between the output sequence and the ground truth retrieved knowledge SEL sequence.
	%
	%Since MetaRetriever aims to retrieve task-specific knowledge based on the input text, we set the ground truth linearized structure sequence as the golden retrieval answer.
	%
	Then, we compute the extraction loss:
	\begin{equation}
		\small
		\mathcal{L}_\text{ext} = \verb|CrossEntropy| \left(f_\theta(x^{\mathcal{S}_i}, k^{\mathcal{S}_i}), y^{\mathcal{S}_i} \right)
	\end{equation}
	where MetaRetriever takes the retrieved knowledge $k^{\mathcal{S}_i}$ with original input and generates the final predicted linearized SEL expression.
	We compute the cross-entropy loss between the predicted linearized SEL and ground truth as the extraction loss.
	Thus, the total inner loss is calculated as follows:
	\begin{equation}
		\small
		\mathcal{L}_\text{inner} = \mathcal{L}_\text{retrv} + \mathcal{L}_\text{ext}
	\end{equation}
	
	The inner loop may consist of several update steps, resulting in multiple updates of the intermediate parameters. Let the number of steps in the inner loop be denoted as $J$, then the model parameters $\theta$ are updated iteratively to $\theta_J$ based on the inner loss $\mathcal{L}_\text{inner}$:
	\begin{equation}
		\small
		\theta_{j+1} = \theta_{j} - \alpha \nabla_{\theta_{j}}  \mathcal{L}_\text{inner}( f_{\theta_{j}} )
	\end{equation}
	where $\alpha$ is the inner step size.

	\paragraph{Outer Loop}
	% In the outer loop, the model parameters are updated by optimizing the performance of $f_{\theta_{J}}$ with respect to $\theta$. As the outer loop simulates the downstream evaluation process, the retrieval loss and the generation loss are calculated using the query set $\mathcal{Q}_{i} = (x^{\mathcal{Q}_i}, y^{\mathcal{Q}_i})$ to mimic the retrieve-then-extract procedure:
	%
	In the outer loop, the model parameters are updated by optimizing the performance of $f_{\theta_{J}}$ with respect to $\theta$. 
	As the outer loop simulates the downstream evaluation process, given the query set $\mathcal{Q}_{i} = (x^{\mathcal{Q}_i}, y^{\mathcal{Q}_i})$, the retrieval loss is calculated to evaluate the retrieval performance:
	\begin{equation}
		\small
		\mathcal{L}_\text{retrv} = \verb|CrossEntropy| \left(f_{\theta_{J}}(x^{\mathcal{Q}_i}), y^{\mathcal{Q}_i} \right)
	\end{equation}
	The extraction loss is computed to assess the extraction results based on the retrieved knowledge:
	\begin{equation}
		\small
		\mathcal{L}_\text{ext} = \verb|CrossEntropy| \left(f_{\theta_{J}}(x^{\mathcal{Q}_i}, k^{\mathcal{Q}_i}), y^{\mathcal{Q}_i} \right)
	\end{equation}
	Additionally, to maintain the ability to understand language, we include a T5-style language modeling loss $\mathcal{L}_{\text{LM}}$~\citep{DBLP:journals/jmlr/RaffelSRLNMZLL20}. And to acquire the capability of generating valid linearized SEL structures, we include a record loss $\mathcal{L}_\text{record}$ as proposed in UIE~\citep{uie2022lu}. 
	The overall objective in the outer loop is calculated using these four losses:
	\begin{equation}
		\small
		\mathcal{L}_\text{outer} = \mathcal{L}_\text{retrv} + \mathcal{L}_\text{ext} + \mathcal{L}_\text{LM} + \mathcal{L}_\text{record}
	\end{equation}
	It is worth noting that the language modeling loss and the record loss are not the primary learning objectives in the downstream IE tasks, thus these losses are calculated based on the original parameters $\theta$ rather than the intermediate parameters $\theta_{J}$. 
	
	Eventually, the model parameters $\theta$ are updated as follows:
	\begin{equation}
		\label{eq:metaupdate}
		\small
		\theta \leftarrow \theta - \beta \nabla_\theta \sum_{\mathcal{T}_i \in \mathcal{T}}  \mathcal{L}_{\text{outer}} ( f_{\theta_{J}})
	\end{equation}
	% where $\beta$ is the outer step size.
	% %
	% Note that the bi-level optimization is performed over the model parameters $\theta$, whereas the objective is computed using the updated model parameters $\theta_{J}$, which will calculate high-order gradients to update parameters.
	% %
	% In effect, our proposed method aims to optimize the model parameters such that one or a small number of gradient steps on a new task will produce maximally effective behavior on that task.
	where $\beta$ is the outer step size. The bi-level optimization is applied to the model parameters $\theta$, but the objective is computed using the updated model parameters $\theta_{J}$, which calculates high-order gradients to update parameters. 
	Essentially, our proposed method optimizes the model parameters such that one or a small number of gradient steps on a new task will yield optimal performance on that task.

	\section{Experiments}
	
	%To verify the effectiveness of our proposed Learn2Refine, we conducted experiments on different IE tasks and settings.
	
	%\subsection{Experimental Settings}
	
	\subsection{Pretraining Datasets}
	
	To pretrain MetaRetriever, we collect a large-scale corpus using CROCODILE\citep{DBLP:conf/emnlp/CabotN21}, an automatic relation extraction dataset construction tool. CROCODILE is based on the distant supervision technique~\citep{DBLP:conf/acl/MintzBSJ09} which aligns texts from English Wikipedia~\footnote{https://www.wikipedia.org/} and knowledge base from Wikidata~\footnote{https://www.wikidata.org/} to create distant labeled relation extraction data. 
	Since distant supervision inevitably introduces noise~\citep{DBLP:conf/acl/MintzBSJ09,DBLP:conf/emnlp/CabotN21}, CROCODILE further uses Natural Language Inference (NLI) to filter data with low confidence. 
	Finally, we construct a 71 million distant labeled pretraining corpus and obtained 6.9 million high-quality pretraining corpus after NLI filtering.
	Detailed statistics of our pretraining corpus are listed in Appendix~\ref{sec:pretraining_corpus}.
	
	Our analysis experiment (Section~\ref{sec:effect_corpus}) indicates that the pretraining corpus we constructed can yield comparable results in the trained model as the data constructed by UIE. 
	As UIE's pretraining data is not open-sourced, \textbf{our data will be the first 10-million-level corpus to be open-sourced, and will be released after the acceptance of the paper.}

	%
	%Our constructed large-scale pretraining corpus will be public in the future.
	%
	% Both versions of our pretraining corpus will be public in the future.

	% To pretrain MetaRetriever, we collect a large-scale pretraining corpus.
	% %
	% We use crocodile~\citep{DBLP:conf/emnlp/CabotN21}, a automatic relation extraction dataset construction tool, to build our pretraining corpus.
	% %
	% Based on distant supervision technique~\citep{DBLP:conf/acl/MintzBSJ09}, crocodile align Wikipedia~\footnote{https://www.wikipedia.org/} texts and Wikidata~\footnote{https://www.wikidata.org/} knowledge base to create distant labeled relation extraction dataset.
	% %
	% Since distant supervision will introduce noise unavoidably~\citep{DBLP:conf/acl/MintzBSJ09,DBLP:conf/emnlp/CabotN21}, it uses Natural Language Inference~(NLI) to filter data with low confidence.
	% %
	% Finally, we construct 71M original distant labeled pretraining corpus and obtain 6.9M high-quality pretraining corpus with NLI filtering.
	%
	%Finally, we construct the distant labeled pretraining corpus with 71M instances.
	%
	%Finally, we construct 6.9M high-quality pretraining corpus.
	%
	% Detailed statistics of our pretraining corpus are listed in Appendix~\ref{sec:pretraining_corpus}.
	%
	%Our constructed large-scale pretraining corpus will be public in the future.
	%
	% Both versions of our pretraining corpus will be public in the future.
	
	\subsection{Evaluation Datasets}
	
	Following previous work~\citep{uie2022lu}, we conduct experiments across 4 IE tasks: named entity recognization (NER), relational triple extraction (RTE), event extraction (EE), structured sentiment extraction (Senti).
	Specifically, we use 12 IE benchmarks in total for these 4 IE tasks\footnote{As NYT datasets overlaps with pre-training data, we didn’t conduct on it for fair comparsion.}: ACE04~\citep{ace2004-annotation}, ACE05~\citep{ace2005-annotation}; CoNLL03~\citep{tjongkimsang2003conll}, CoNLL04~\citep{roth-yih-2004-linear}, SciERC~\citep{luan-etal-2018-multi}, CASIE~\citep{Satyapanich_Ferraro_Finin_2020}, SemEval-14~\citep{pontiki-etal-2014-semeval}, SemEval-15~\citep{pontiki-etal-2015-semeval},
	SemEval-16~\citep{pontiki-etal-2016-semeval}.
	The detailed statistics of these datasets can be seen in Appendix~\ref{sec:dataset_stat}.

	\subsection{Evaluation Metrics}

	\begin{table}[!t]
		\centering
		\resizebox{1.0\linewidth}{!}{
			\begin{tabular}{cc|c|c}
				\toprule
				\textbf{Task} & \textbf{Dataset} & \textbf{UIE} & \textbf{MetaRetriever} \\
				
				\midrule
				\multirow{3}{*}{NER} 
				& ACE04 & 85.69 & \textbf{86.10} {\small (0.41$\uparrow$)}\\
				& ACE05-Ent & 83.88 & \textbf{84.01} {\small (0.13$\uparrow$)}\\
				& CoNLL03 & 91.94 & \textbf{92.38} {\small (0.44$\uparrow$)}\\
				
				\midrule
				\multirow{3}{*}{RTE} 
				& ACE05-Rel & 62.73 & \textbf{64.37} {\small (1.64$\uparrow$)}\\
				& CoNLL04 & 73.48 & \textbf{73.66} {\small (0.18$\uparrow$)}\\
				& SciERC & 35.35 & \textbf{35.77} {\small (0.42$\uparrow$)}\\
				
				\midrule
				\multirow{2}{*}{EE Trg.} 
				& \multirow{1}{*}{ACE05-Evt} & 71.33 & \textbf{72.38} {\small (1.05$\uparrow$)}\\
				& \multirow{1}{*}{CASIE} & 69.14 & \textbf{69.76} {\small (0.62$\uparrow$)}\\
				
				\midrule
				\multirow{2}{*}{EE Arg.} 
				& \multirow{1}{*}{ACE05-Evt} & 50.62 & \textbf{52.62} {\small (2.00$\uparrow$)}\\
				& \multirow{1}{*}{CASIE}  & 58.56  & \textbf{60.37} {\small (1.81$\uparrow$)}\\
				
				\midrule
				\multirow{4}{*}{Senti} 
				& 14-res & 72.86 & \textbf{73.41} {\small (0.55$\uparrow$)}\\
				& 14-lap & 62.68 & \textbf{62.83} {\small (0.15$\uparrow$)}\\
				& 15-res & 65.51 & \textbf{65.85} {\small (0.34$\uparrow$)}\\
				& 16-res & 73.26 & \textbf{73.55} {\small (0.29$\uparrow$)}\\
				\bottomrule
			\end{tabular}
		}
		\caption{Overall results on 12 datasets in the fully supervised settings. The performance of UIE are reported based on the official UIE-en-base model.}
		\label{tab:fully_supervised}
		
	\end{table}

	We use span-based offset Micro-F1 as the primary metric to evaluate the model for different IE tasks:

	\begin{table*}[!t]
		\centering
		
		\begin{tabular}{cl|ccc|ccc}
			\toprule
			\multirow{2}{*}{\textbf{Task}} & \multirow{2}{*}{\textbf{Model}} & \multicolumn{3}{c|}{\textbf{Few-Shot}} &  \multicolumn{3}{c}{\textbf{Low-Resource}} \\
			& & \textbf{1-Shot} & \textbf{5-Shot} & \textbf{10-Shot}  & \textbf{1\%} & \textbf{5\%} & \textbf{10\%} \\
			
			\midrule
			\multicolumn{1}{c}{\multirow{2}[2]{*}{\shortstack{\textbf{NER} \\ (\textbf{CoNLL03})}}} 
			& UIE & 46.43 & 67.09 & 73.90 & 82.84 & 88.34 & 89.63  \\
			& MetaRetriever & \textbf{49.44} & \textbf{69.88} & \textbf{74.19} & \textbf{83.45} & \textbf{89.11} & \textbf{90.42} \\
			& $\Delta$ & 3.01$\uparrow$ & 2.79$\uparrow$ & 0.29$\uparrow$ & 0.61$\uparrow$ & 0.77$\uparrow$ & 0.79$\uparrow$ \\

			\midrule
			\multicolumn{1}{c}{\multirow{2}[2]{*}{\shortstack{\textbf{RTE} \\ (\textbf{CoNLL04})}}} 
			& UIE & 22.05 & 45.41 & 52.39 & 30.77 & 51.72 & 59.18 \\
			& MetaRetriever & \textbf{29.90} & \textbf{47.02} & \textbf{53.95} & \textbf{36.31} & \textbf{52.59} & \textbf{59.45} \\
			& $\Delta$ & 7.85$\uparrow$ & 1.61$\uparrow$ & 1.56$\uparrow$ & 5.54$\uparrow$ & 0.87$\uparrow$ & 0.27$\uparrow$ \\
			
			\midrule
			\multicolumn{1}{c}{\multirow{2}[2]{*}{\shortstack{\textbf{Event Trigger} \\ (\textbf{ACE05-Evt})}}} 
			& UIE & 38.14 & \textbf{51.21} & 53.23 & 41.53 & 55.70 & 60.29 \\
			& MetaRetriever & \textbf{39.85} & 49.43 & \textbf{53.58} & \textbf{43.98} & \textbf{58.35} & \textbf{63.19} \\
			& $\Delta$ & 1.71$\uparrow$ & 1.78$\downarrow$ & 0.35$\uparrow$ & 2.45$\uparrow$ & 2.65$\uparrow$ & 2.90$\uparrow$ \\

			\midrule
			\multicolumn{1}{c}{\multirow{2}[2]{*}{\shortstack{\textbf{Event Argument} \\ (\textbf{ACE05-Evt})}}} 
			& UIE & 11.88 & 27.44 & \textbf{33.64} & 12.80 & 30.43 & 36.28 \\
			& MetaRetriever & \textbf{13.30} & \textbf{27.70} & 32.31 & \textbf{14.86} & \textbf{31.85} & \textbf{37.85} \\
			& $\Delta$ & 1.42$\uparrow$ & 0.26$\uparrow$ & 1.33$\downarrow$ & 2.06$\uparrow$ & 1.42$\uparrow$ & 1.57$\uparrow$ \\
			
			\midrule
			\multicolumn{1}{c}{\multirow{2}[2]{*}{\shortstack{\textbf{Sentiment} \\ (\textbf{16res})}}}
			& UIE & 10.50 & 26.24 & 39.11 & 24.24 & 49.31 & 57.61 \\
			& MetaRetriever & \textbf{18.80} & \textbf{34.14} & \textbf{43.53} & \textbf{34.47} & \textbf{51.78} & \textbf{58.79} \\
			& $\Delta$ & 8.30$\uparrow$ & 7.90$\uparrow$ & 4.42$\uparrow$ & 10.23$\uparrow$ & 2.47$\uparrow$ & 1.18$\uparrow$ \\
			
			\bottomrule
		\end{tabular}%
		\caption{Few-Shot and Low-resource results. Results of UIE are reported from original paper~\citep{uie2022lu}.}
		\label{tab:fslr}
		
	\end{table*}

	(1) \textbf{NER}: an entity is correct if its offsets and type are correct.
	
	(2) \textbf{RTE}: a relation is correct if its relation type is correct and the offsets and entity types of the related head/tail entity are correct.
	
	(3) \textbf{EE Trg.}: an event trigger is correct if its offsets and event type are correct.
	
	(4) \textbf{EE Arg.}: an event argument is correct if its offsets, role type, and event type are correct.
	
	(5) \textbf{Senti}: a sentiment triple is correct if the offsets boundary of the target, the offsets boundary of the opinion span, and the target sentiment polarity are correct.

	After generation, we reconstruct the offset of predicted information pieces by finding the matched utterance in the input sequence one by one.
	
	To validate the effectiveness of our proposed method, we compare MetaRetriever with the state-of-the-art model UIE~\citep{uie2022lu} in fully-supervised, few-shot and low-resource scenarios.
	
	\subsection{Main Results}

	\tablename~\ref{tab:fully_supervised} and \tablename~\ref{tab:fslr} summarize the results of our proposed MetaRetriever compared to the previous work, UIE, on 4 types of IE tasks. 
	From the experimental results, we have several observations
	(1) \textbf{Equipped with task-specific knowledge, MetaRetriever outperforms UIE on all fully-supervised, few-shot, and low-resource scenarios.} MetaRetriever surpasses UIE with an average of 0.7\%, 2.5\%, and 2.3\% in F1 score respectively in the fully-supervised, few-shot, and low-resource settings. This strongly demonstrates that MetaRetriever can utilize the retrieved knowledge to improve IE performance. 
	(2) \textbf{Empowered by meta-pretraining, MetaRetriever can effectively learn downstream task-specific extraction targets.} Despite the fact that event extraction and sentiment extraction were not present in the pretraining phase, MetaRetriever still outperforms UIE in these tasks, validating that MetaRetriever can quickly learn downstream IE tasks through the meta-pretraining algorithm to support task-specific extraction. 
	(3) \textbf{Benefiting from task-specific knowledge and meta-pretraining, MetaRetriever exhibits significant superior performance in data-scarce scenarios.}
	MetaRetriever achieves large gains in the few-shot and low-resource settings. 
	This is attributed to two factors: first, the meta-pretraining algorithm can quickly learn downstream IE tasks, which reduces the requirement for training data, and second, the retrieved task-specific knowledge provides crucial clues for the model to make correct predictions, thus MetaRetriever can achieve impressive performance in data-scarce scenarios.
	
	\section{Analysis}
	
	\subsection{Effect of Meta-Pretraining Algorithm}
	
		\begin{table}[!th]
		\centering
		\resizebox{1.0\linewidth}{!}{
			\begin{tabular}{l|ccc|ccc}
				\toprule
				& \multicolumn{3}{c|}{CoNLL03} & \multicolumn{3}{c}{CoNLL04} \\
				& Sup. & Few. & Low. & Sup. & Few. & Low. \\
				\midrule
				MetaRetriever & 92.38 & 49.44 & 83.45 & 73.66 & 29.90 & 36.31 \\
				SimpleRetriever & 92.14 & 43.72 & 82.66 & 73.33 & 24.47 & 32.43 \\ 
				%		w/o knowledge & 91.96 & 71.60 \\
				\bottomrule
			\end{tabular}%
		}
		
		\caption{Effect of Meta-Pretraining Algorithm.}
		\label{tab:effect_meta}
	\end{table}
	
	To examine the contribution of our Meta-Pretraining Algorithm, we remove the inner loop in Algorithm~\ref{alg:pretrain} and pretrain our model using only the outer loop. 
	Without the inner loop, the bi-level optimization is reduced to single-level optimization. 
	We name this pretrained model SimpleRetriever and compare its performance with MetaRetriever on CoNLL03 and CoNLL04 datasets in fully-supervised, few-shot(1-shot) and low-resource(1\%) settings.
	The results are presented in Table~\ref{tab:effect_meta} and it can be observed that: 
	Compared to MetaRetriever, the performance of SimpleRetriever declines on all datasets in all settings. Notably, in data-scarce scenarios, SimpleRetriever underperforms MetaRetriever by 3.92\% on average. This demonstrates that the meta-pretraining algorithm enables the model to quickly learn downstream IE tasks and thus improve performance.

	\subsection{Effect of Finetune Epoch}
	
	\begin{figure}[!t]
		\centering
		\includegraphics[scale=0.45]{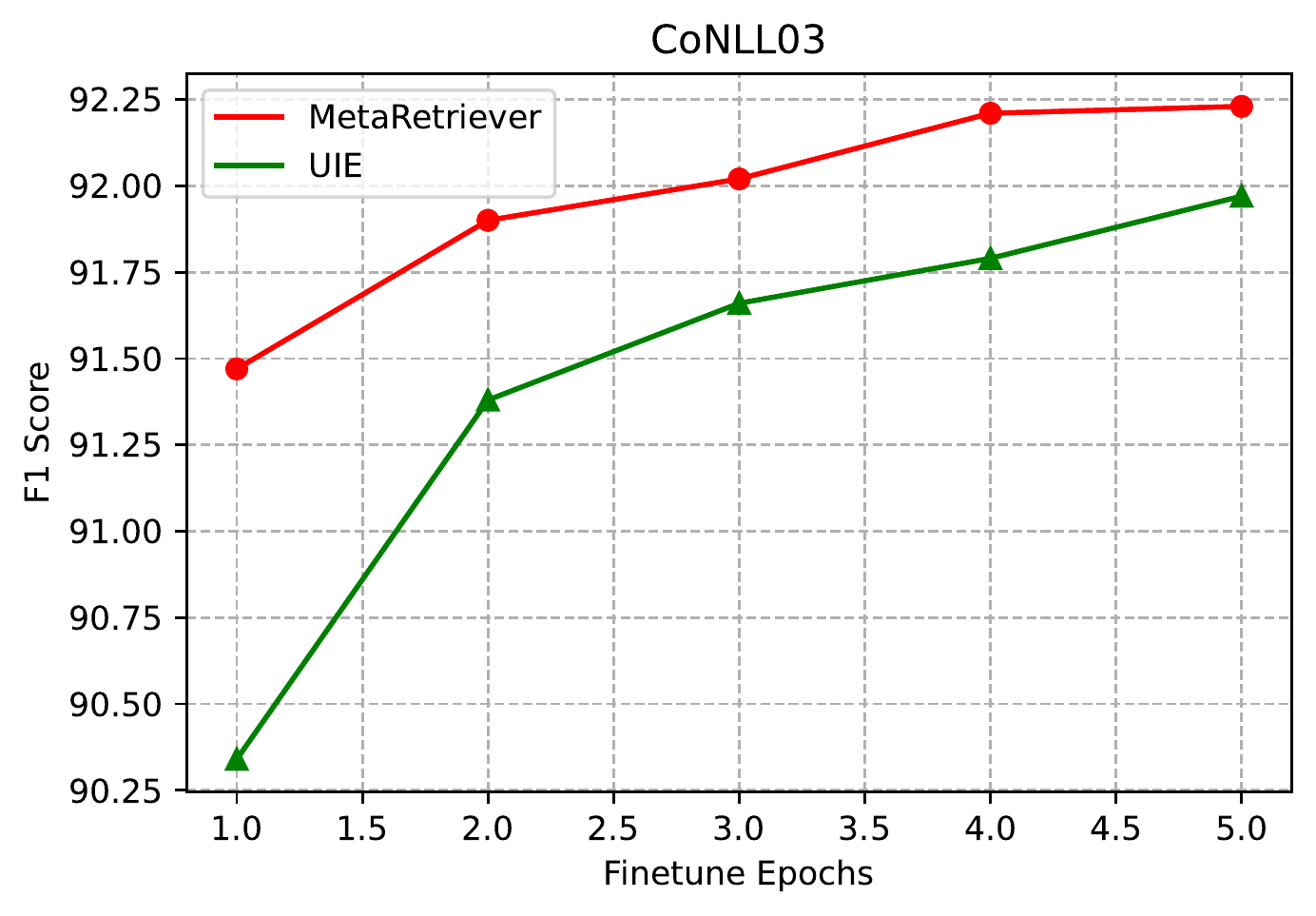}
		\caption{Effect of Finetune Epoch.}
		\label{fig:finetune-epoch}
	\end{figure}
	
	In order to demonstrate the efficiency of our MetaRetriever in rapidly learning downstream information extraction tasks, we conducted a series of experiments. 
	Specifically, we compare the performance of MetaRetriever to that of UIE across a range of finetuning epochs, from 1 to 5 on CoNLL03 dataset. 
	The results of these experiments, presented in Figure~\ref{fig:finetune-epoch}, clearly indicate that MetaRetriever outperforms UIE from the very first finetuning epoch. 
	Furthermore, as the number of finetuning epochs increases, the performance of MetaRetriever consistently surpasses that of UIE, with the greatest difference observed when comparing MetaRetriever's performance in 2 finetuning epochs to UIE's performance in 5 finetuning epochs. 
	These results strongly suggest that MetaRetriever's ability to rapidly learn downstream tasks is significantly enhanced by its use of meta-pretraining. Additionally, we also conducted experiments on the CoNLL04 datasets, the results of which are provided in Appendix~\ref{sec:details_effect_finetune_epoch}.

	\subsection{Effect of Structure Complexity}
	\label{sec:effect_corpus}
	
	\begin{figure}[h]
		\centering
		\includegraphics[scale=0.45]{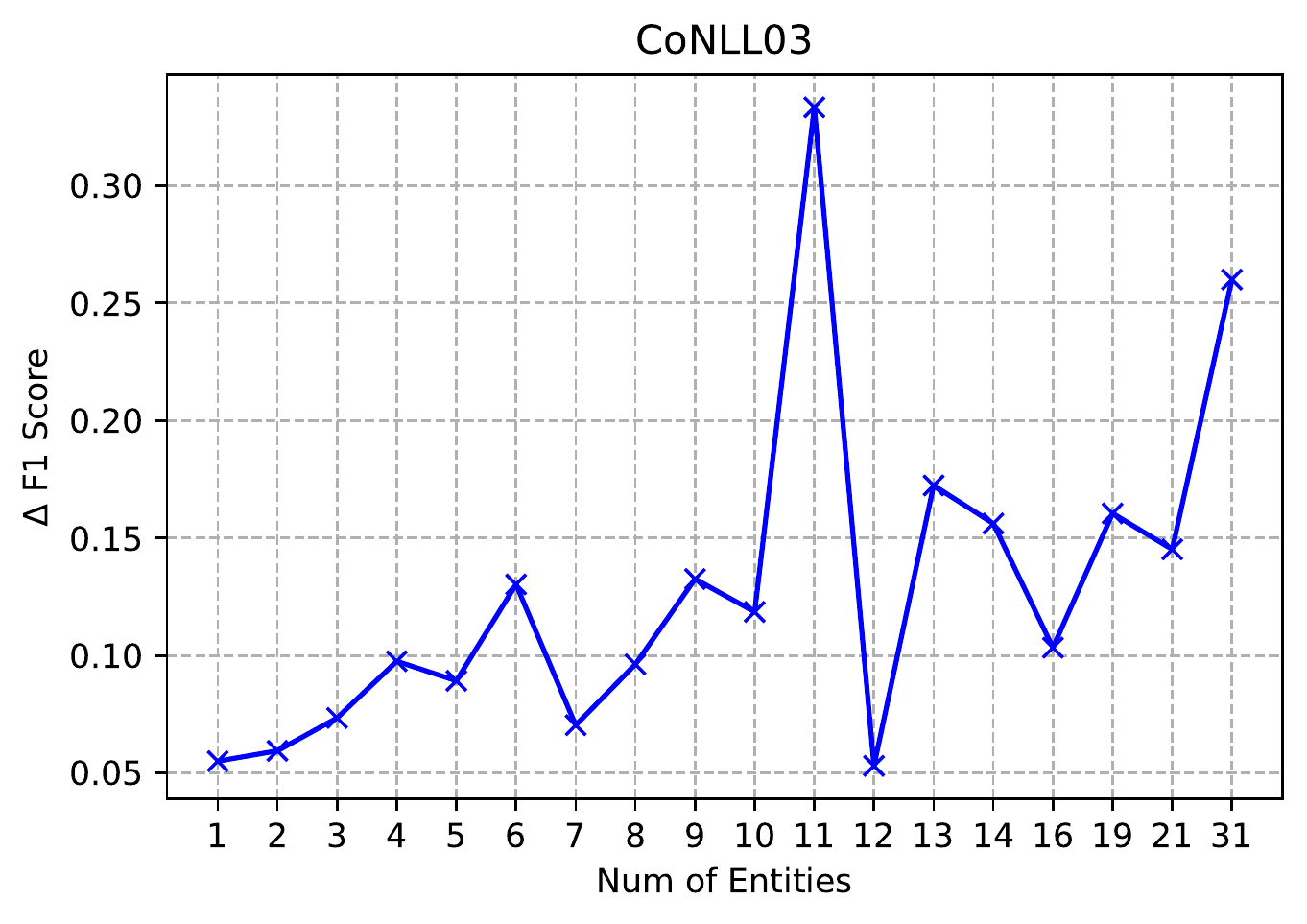}
		\label{fig:complicated-structure-conll03}
		\caption{Effect of Structure Complexity.}
		\label{fig:complicated-structure}
	\end{figure}
	
	\noindent In order to determine the effectiveness of MetaRetriever in dealing with complex structures when compared to UIE, we conduct an experiment using the test set of CoNLL03, grouping the test instances according to the number of entities present. 
	We report the performance gain of MetaRetriever compared with UIE in different group in \figurename~\ref{fig:complicated-structure}.
	We can obverse that, as the number of entities present in a test instance increases, MetaRetriever's performance advantage over UIE also increases. 
	This demonstrates that MetaRetriever is able to effectively leverage task-specific knowledge to handle complex structures and improve information extraction performance. 
	We also conduct further experiments on the CoNLL04 dataset, the results of which can be found in Appendix~\ref{sec:details_effect_complicated_structure}.
	
	\subsection{Effect of Pretraining Corpus Collection}

	\begin{table}[!th]
		\centering
		%	\resizebox{1.0\linewidth}{!}{
			\begin{tabular}{c|c|c}
				\toprule
				& UIE & UIE-ours \\
				\midrule
				\multirow{1}*{CoNLL03} & 91.94 & 91.96 \\
				\multirow{1}*{CoNLL04} & 73.48 & 71.60 \\
				\multirow{1}*{ACE05 Trg.} & 71.33 & 73.39 \\
				\multirow{1}*{ACE05 Arg.} & 50.62 & 51.87 \\
				\multirow{1}*{16res} & 73.26 & 72.16 \\
				\midrule
				\multicolumn{1}{c|}{AVG} & 72.13 & 72.20 \\ % 0.07
				
				\bottomrule
			\end{tabular}%
			%	}
		
		\caption{Effect of Pretraining Corpus Collection.}
		\label{tab:effect_corpus}
	\end{table}

	\begin{table*}[!t]
		\centering
		\resizebox{0.95\textwidth}{!}{
			\begin{tabular}{ll}
				\toprule
				\multicolumn{2}{l}{(1) \textit{Gale-force winds were reported in California late Wednesday, where gusts reached 69 mph at Ontario and 57 mph at El Toro.}} \\
				\midrule
				UIE & \texttt{((LOC: California)(LOC: Ontario)(LOC: El Toro))} \\
				MetaRetriever & \texttt{((LOC: California)(LOC: Ontario\textcolor{red}{(Located\_In: California)})(LOC: El Toro\textcolor{red}{(Located\_In: California)}))}\\
				\textsc{Retrieved Knoledge} & \texttt{((LOC: California)(LOC: Ontario)(LOC: El Toro(Located\_In: California)))}\\
				
				\midrule
				\multicolumn{2}{l}{\textit{(2) Alekseev, along with others in the Soviet diplomatic corps, were not receiving accurate information from Moscow.}} \\
				\midrule 
				UIE & \texttt{(PER: Alekseev (Live\_In: Soviet)) (LOC: Soviet) (LOC: Moscow))}  \\
				MetaRetriever & \texttt{(PER: Alekseev (Live\_In: Soviet)) (LOC: Soviet) (LOC: Moscow\textcolor{red}{(Located\_In: Soviet)}))} \\
				\textsc{Retrieved Knoledge} & \texttt{(PER: Alekseev (Live\_In: Soviet)) (LOC: Soviet) (LOC: Moscow(Located\_In: Soviet)))}\\
				
				\bottomrule
			\end{tabular}
		}
		\caption{Case Study.}
		\label{tab:case_study}
	\end{table*}

	To evaluate the quality of the corpus we have collected for pretraining, we use this corpus to pretrain a variant of UIE which we refer to as UIE-ours. 
	We then compared UIE-ours to the original UIE on four representative datasets for each IE task in fully-supervised settings.
	The results, presented in Table~\ref{tab:effect_corpus}, show that UIE-ours, which is pretrained on our collected corpus, is comparable in performance to the original UIE (with an average gain of 0.07\%). 
	This indicates that our collected corpus is of similar quality to that used for training the original UIE and we plan to make it publicly available to promote future research. 
	Additionally, this also demonstrates that the improvement of MetaRetriever is due to our proposed method rather than the corpus used for pretraining, further validating the effectiveness of our proposed method.

	\subsection{Effect of Support-Query Construction for Meta-Pretraining}
	\label{sec:exp_graph_match}
	
	Conventional meta-learning methods typically employ the episodic training technique to construct "tasks" for meta-training~\citep{DBLP:conf/nips/VinyalsBLKW16,DBLP:conf/icml/FinnAL17}.
	This technique creates "tasks" by randomly sampling instances based on their category, which requires random access to all data.
	However, since large-scale corpora cannot be stored in memory and must be accessed from external storage instead, random-sampling operation will cause long IO time. 
	To address this, we design a support-query pairing method which can significantly reduce IO times. 
	To illustrate the efficiency of our method, we compared the time cost of our method with episodic training on 10K instances.
	Episodic training required 100.42s to construct "IE tasks" while our method only required 5.57s, demonstrating the significant efficiency of our method.
	
	\subsection{Case Study}

	In this study, we compare the performance of our MetaRetriever model with that of the UIE model using a selection of cases from the CoNLL04 dataset, as illustrated in Table~\ref{tab:case_study}. 
	
	In the first case, our MetaRetriever, aided by the retrieved task-specific knowledge, ``\texttt{(LOC: El Toro (Located\_In: California))}'', correctly predict ``\texttt{(LOC: Ontario (Located\_In: California))}''.
	In contrast, the UIE model was unable to make a precise prediction without the benefit of this task-specific knowledge. This demonstrates that models can significantly benefit from the incorporation of retrieved task-specific knowledge to generate more accurate linearized SEL expressions.
	
	In the second case, we observe that MetaRetriever makes a redundant prediction, ``\texttt{(LOC: Moscow (Located\_In: Soviet))}''.
	Though this relational triple is not labeled in the CoNLL04 dataset, it is a common sense knowledge that Moscow is located in Russia, which is former Soviet Union. MetaRetriever was able to predict this based on the background knowledge acquired from the pretraining corpus.
	
	\section{Related Work}
	
	Information Extraction~(IE) aims at extracting text spans or a tuple of text spans of interests from plain texts.
	There exist many specific IE tasks: Named Entity Recognition~\citep{lample-etal-2016-neural,lin-etal-2019-sequence}, Relational Triple Extraction~\citep{zheng-etal-2017-joint,levy-etal-2017-zero}, Event Extraction~\citep{lin-etal-2018-nugget,wadden-etal-2019-entity,du-cardie-2020-event,DBLP:journals/taslp/LiPLWNWYW22,DBLP:journals/corr/abs-2211-08168,li2022survey}, etc.
	%
	%To achieve these different IE tasks, researchers have proposed many paradigm, such as sequence labeling~\citep{lample-etal-2016-neural,zheng-etal-2017-joint,lin-etal-2019-sequence}, span classification~\citep{sohrab-miwa-2018-deep,lin-etal-2018-nugget,wadden-etal-2019-entity}, and MRC~\citep{levy-etal-2017-zero,li-etal-2020-unified,du-cardie-2020-event}.
	%
	For a long time, researchers are devoted to propose task-customized and isolated methods to achieve these different IE tasks.
	%
	%Several task-specific pretraining techniques are also proposed~\citep{mengge-etal-2020-coarse,wang-etal-2021-cleve, qin-etal-2021-erica}.
	%
	%In the recent years, with the rising of pretraining technique, pretraining a universal model for several NLP tasks has attracted a lot of attention, e.g., contextualized representation \citep{devlin-etal-2019-bert, roberta}, text generation \citep{lewis-etal-2020-bart,2020t5}, multi-modal \citep{li-etal-2021-unimo,pmlr-v139-cho21a}, and multi-lingual \citep{conneau-etal-2020-unsupervised,xue-etal-2021-mt5}.
	%
	In the recent years, with the rising of pretraining technique, pretraining a universal model for several NLP tasks has attracted a lot of attention~\citep{devlin-etal-2019-bert,roberta,lewis-etal-2020-bart,2020t5,xue-etal-2021-mt5}.
	Following this trend, several works make attempts to unify diverse IE tasks.
	\citet{DBLP:conf/acl/YanGDGZQ20} first propose a span-offset-based generation to solve various NER tasks in a universal manner.
	\citet{DBLP:conf/acl/YanDJQ020} use a similar way to solve different aspect-based sentiment analysis tasks.
	%
	%\citep{uie2022lu} proposes UIE which designs Structured Extraction Language to formulate all IE tasks in a unified form and pretrains a encoder-decoder Transformer to generate SEL to achieve information extraction.
	%
	%However, SEL will become complicated when there exists multiple information to be extracted, causing poor performance.
	%
	\citet{uie2022lu} propose UIE which designs Structured Extraction Language to formulate all IE tasks in a unified form but it will experience poor performance when there exists complex structure to be extracted.
	%
	%Existing works~\citep{DBLP:conf/acl/WangJBWHHT20,DBLP:journals/corr/abs-2202-09022} have shown that retrieving knowledge from external resources can provide important clues to make accurate extraction but suffer from time-consuming resource construction and sub-optimal heuristic retrieval strategy design.
	%
	Several works have shown that retrieving knowledge from external knowledge bases can empower models to generate complex sequences~\citep{lewis2020retrieval,DBLP:conf/acl/WangJBWHHT20,cai2022recent,DBLP:journals/corr/abs-2202-09022} and a large number of knowledge stored in pretrained language models can be ``retrieved'' explicitly~\citep{DBLP:conf/emnlp/PetroniRRLBWM19,DBLP:conf/emnlp/RobertsRS20,zhang2022automatic,yu2022generate}.
	In this paper, we propose MetaRetriever to retrieve task-specific knowledge from PLMs to enhance universal IE.
	
	Meta-learning aims to learn better learning algorithms from a series of tasks, i.e., learning to learn.
	In meta-learning, there exist diverse branches focusing on different aspects of the learning algorithm: learning-to-initialize~\citep{DBLP:conf/icml/FinnAL17,DBLP:conf/nips/FinnXL18}, learning-to-compare~\citep{DBLP:conf/nips/VinyalsBLKW16,DBLP:conf/nips/SnellSZ17}, learning-to-optimize~\citep{DBLP:conf/iclr/RaviL17}, etc.
	%
	% Our work is inspired by MAML~\citep{DBLP:conf/icml/FinnAL17}, a learning-to-initialize method.
	%
	Existing works are most devoted to solve the few-shot problem, cross-domain problem, etc.
	We make the first attempt to adapt meta-learning into universal IE.
	% 可能有风险
	
	\section{Conclusion}
	
	% In this paper, we propose MetaRetriever which can retrieve task-specific knowledge from itself to enhance universal information extraction.
	% To make it task-specific knowledge and ignore noise knowledge, we further propose a meta-pretraining algorithm to make MetaRetriever can fast learn downstream IE tasks.
	% Experimental results demonstrate that MetaRetriever achieve the new state-of-the-art on fully-supervised, low-resource, and few-shot scenarios on 4 IE tasks, 12 datasets.
	% Further analysis experiments also validate the effectiveness and contributions of our MetaRetriever.
	
	In this work, we present MetaRetriever, a novel approach for enhancing universal information extraction through the retrieval of task-specific knowledge from within the model itself. 
	To ensure that the retrieved knowledge is truly task-specific and to minimize the impact of irrelevant or "noise" knowledge, we also propose a meta-pretraining algorithm that enables MetaRetriever to quickly adapt to downstream information extraction tasks. 
	Empirical results demonstrate the effectiveness of our approach, with MetaRetriever achieving new state-of-the-art performance on a range of information extraction tasks across 12 datasets in fully-supervised, low-resource, and few-shot scenarios. 
	Further analysis experiments also validate the contributions and effectiveness of MetaRetriever.
	
	\section{Limitations}
	
	While our MetaRetriever has demonstrated its superior performance on 4 IE tasks, 12 datasets, in fully-supervised, few-shot and low-resource scenarios, it still has several limitations.
	First, since MetaRetriever will first retrieve task-specific knowledge and then make predictions in the inference phase, such a retrieve-then-extract manner will take longer inference time than non-retrieve methods unavoidably.
	%
	%It will cost about twice time as previous work to make a prediction.
	%
	Second, our proposed meta-pretraining algorithm is based on bi-level optimization.
	In the pretraining phase, it needs to calculate high-order gradients to optimize parameters and calculating high-order gradient requires more time, which will cause longer time to pretrain MetaRetriever.
	%
	% Therefore, it takes longer time to pretrain MetaRetriever.
	%
	Detailed discussions are shown in Appendix~\ref{sec:efficiency}.

	\section{Ethical Considerations}
	
	This paper constructs a new dataset for universal information extraction, and we discuss some related ethical considerations here. 
	(1) \textbf{Intellectual property}. The Wikipedia corpus is shared under the CC BY-SA 3.0 license\footnote{\url{https://creativecommons.org/licenses/by-sa/3.0}} and Wikidata is shared under the CC0 1.0 license\footnote{https://creativecommons.org/publicdomain/zero/1.0/}. 
	CROCODILE is licensed under the CC BY-SA-NC 4.0 license\footnote{https://creativecommons.org/licenses/by-nc-sa/4.0/}.
	Our data source and the construct tool are all free for research use. 
	(2) \textbf{Controlling Potential Risks.} Since the texts in Wikipedia do not involve private information and annotating entities and relations does not require many judgments about social issues, we believe our collected dataset does not create additional risks.
	To ensure it, we manually checked some randomly sampled data and did not note risky issues.
	(3) \textbf{Worker Treatments.} Since the dataset construction procedure is automatic without any manual annotation, it does not create potential risk about the worker treatments. 
	
	\section*{Acknowledgements}
	This work is supported by the National Key Research and Development Program of China (grant No.2021YFB3100600), the Strategic Priority Research Program of Chinese Academy of Sciences (grant No.XDC02040400) and the Youth Innovation Promotion Association of CAS (Grant No. 2021153). 
	
	\bibliography{anthology,custom}
	\bibliographystyle{acl_natbib}
	
	\appendix
	
	\section{Details of Pretraining Corpus}
	\label{sec:pretraining_corpus}
	
	\begin{table}[tbhp]
		\centering
		\begin{tabular}{l|r}
			\toprule
			\# Sentence & 71,947,589 \\
			\# Entity & 208,261,778 \\
			\# Triple & 27,011,527 \\
			%			\# Sent. w/ Entity & 71,947,589 \\
			%			\# Sent. w/ Triple & 6,161,291\\
			\# Entity Types & 45,062 \\
			\# Relation Types & 1,465 \\
			Average Length & 33.43\\
			\bottomrule
		\end{tabular}%
		\caption{Statistics of Raw Pretraining Corpus.}
		\label{tab:corpus_stat_raw}
	\end{table}
	
	\begin{table}[tbhp]
		\centering
		\begin{tabular}{l|r}
			\toprule
			\# Sentence & 6,951,695 \\
			\# Entity & 48,352,090 \\
			\# Triple & 12,497,480 \\
			%		\# Sent. w/ Entity & 6,951,695 \\
			%		\# Sent. w/ Triple & 6,951,695 \\
			\# Entity Types & 39,511 \\
			\# Relation Types & 1,255 \\
			Average Length & 47.47 \\
			\bottomrule
		\end{tabular}%
		\caption{Statistics of Filtered Pretraining Corpus.}
		\label{tab:corpus_stat_filtered}
	\end{table}

	We apply CROCODILE\footnote{https://github.com/Babelscape/crocodile}~\citep{DBLP:conf/emnlp/CabotN21}, which is a automatic distant supervised labeling tool developed for relation extraction, to collect our pretraining corpus.
	CROCODILE utilizes widely-used distant supervision technique~\citep{DBLP:conf/acl/MintzBSJ09} to build relation extraction dataset by aligning English Wikipedia\footnote{https://www.wikipedia.org/} texts and Wikidata\footnote{https://www.wikidata.org/} knowledge base.
	Additionally, CROCODILE applies RoBERTa NLI model~\footnote{\href{https://huggingface.co/joeddav/xlm-roberta-large-xnli}{xlm-roberta-large-xnli}} to filter distant labeled data with low-confidence.
	Finally, we construct 71M original pretraining corpus without NLI filtering and obtain 6.9M high-quality pretraining corpus with NLI filtering.
	The statistics of them are shown in \tablename~\ref{tab:corpus_stat_raw} and \tablename~\ref{tab:corpus_stat_filtered} respectively.
	Since our collected corpus contains the 10-million-level sentences covering about 40K entity types and over 1K relation types, we believe the collected corpus is diverse enough in genre and topic, which exhibits a good coverage for general domain . 
	Both of these pretraining corpus will be public after acceptance.

	\section{Details of Meta-Pretraining Dataset Construction Procedure}
	
	\label{sec:pairing}
	
	The goal of our meta-pretraining is to train MetaRetriever that can quickly learn the extraction target semantics of various downstream IE tasks for retrieval using only a few training epochs.
	To accomplish this, MetaRetriever is pretrained on a set of ``IE tasks'' based on the meta-learning technique~\citep{DBLP:conf/icml/FinnAL17}, such that the trained model can quickly learn new IE tasks using only a small number of trials.
	%
	%In effect, the meta-learning treats entire tasks as training examples.
	%
	%In this section, we detail how we construct various ``IE tasks'' to meta-pretrain our model.
	%
	To construct various ``IE tasks'', a intuitive way is to follow the \textit{episodic training} technique~\citep{DBLP:conf/nips/VinyalsBLKW16} which are widely used in conventional meta-learning methods.
	In \textit{episodic training}, it randomly samples $N$ classes from original class set as a $N$-classification ``task''.
	For each $N$-classification task, it will randomly samples $K$ instances as the support set to mimic the training set for this task and then use remain instances as the query set to mimic the test set, i.e., the $N$-way-$K$-shot setting. 
	In each training epoch, it will sample different $N$-way-$K$-shot tasks to train models.
	However, when pretraining a model, it is time-consuming and memory-consuming to sample classes from a large-scale corpus.
	To meta-pretrain MetaRetriever, we design Support-Query Pairing Algorithm based on the graph maximum-weighted matching algorithm~\citep{DBLP:journals/csur/Galil86}.
	Our Support-Query Pairing Algorithm can be divided into three steps: (1) Partition by class; (2) Deduplication; (3) Graph Matching.
	
	\paragraph{Partition by class}
	Different from \textit{episodic training} which treats each a support set and a query set as a task, we construct a ``task'' with only two instances.
	One acts as the support instance and the other as the query instance, i.e., the size of our support set and query are both 1.
	To make the support instance ``support'' the query instance, their classes should be as nearly identical as possible so the model can learn the semantics of the target classes from the support instance to make predictions on the query instance.
	To achieve this goal, we partition all pretraining instances based on the class.
	Specifically, we denote the raw training dataset $\mathcal{D} = \{ (x_i, y_i) \}_{i=1}^{N}$ where $x_i$ refers to the raw text, $y_i$ refers to the golden linearized SEL expression, and $N$ refers to the number of the ;pretraining instances.
	Given the training corpus $\mathcal{D}$, we can get the class set $\mathbb{C}$ which contains all SpotNames and AssoNames of $\mathcal{D}$.
	Then, we collect instances for each classes $\mathcal{D}_{c}, \ c \in \mathbb{C}$ where $\mathcal{D}_{c}$ refers to all instances which contains $c$-type target information piece.
	Therefore, we can get the partitioned dataset $\mathcal{D}_{\mathbb{C}}$.
	
	\paragraph{Deduplication}
	Getting the partitioned dataset $\mathcal{D}_{\mathbb{C}}$, there exists serious duplication phenomenon because one instance will contain several classes and thus it will be partitioned into multiple subset $\mathcal{D}_{c}$.
	Such a duplication will cause a severe data imbalance since if one instance contains $m$ classes, it will be duplicated $m$ times, hurting the model performance.
	To alleviate this problem, we should deduplicate the partitioned dataset $\mathcal{D}_{\mathbb{C}}$.
	Specifically, we sort the class set $\mathbb{C}$ by the number of instances of each class.
	From less to more, we remove these instances from the large subset which have existed in the small subset.
	After that, we get the deduplicated partitioned dataset $\hat{\mathcal{D}}_{\mathbb{C}}$.

	\paragraph{Graph Matching}
	Since we treat a support-query instance pair as a task, we want to find a optimal pairing solution that the classes of the support instance could cover the classes of the query instance as many as possible globally.
	To achieve this goal, we first define a pairing score function to evaluate a support-query pair.
	Given a support instance $s$ and a query instance $q$, we define the pairing score function as follows,
	\begin{equation}
		\rho(s, q) = \frac{|F_\mathbb{C}(s) \cap F_\mathbb{C}(q)|}{|F_\mathbb{C}(s)|} + \frac{1}{|F_\mathbb{C}(s)|}
	\end{equation}
	where $F_\mathbb{C}(\cdot)$ means the class set contained in the given instance.
	In this score function, the first term evaluates the ratio how the support instance covers the query instance.
	The second term evaluates the number of the classes of the support instance.
	We wish the support instance should cover the query instance exactly so the number of the classes of the support instance should be as small as possible besides the class coverage.
	Then, we define a matching score function to calculate the pairing degree given any two instances $x$ and $y$:
	\begin{equation}
		\varphi(x, y) = \max \left\{ \rho(x,y), \rho(y,x) \right\}
	\end{equation}
	Given a subset $\mathcal{D}_{c}, \ c \in \mathbb{C}$, we can construct a weighted undirected complete graph $\mathcal{G}_{c}$.
	In $\mathcal{G}_{c}$, each instance in $\mathcal{D}_{c}$ acts as a node and the arbitrary two nodes $x$ and $y$ have a edge with $\varphi(x, y)$ weight.
	Next, we run the maximum-weighted graph matching algorithm~\citep{DBLP:journals/csur/Galil86} to get a graph matching result $\mathcal{M}_{c}$.
	In $\mathcal{M}_{c}$, each node matching pair represents a support-query pair.
	We run the aforementioned matching procedure for each class $c \in \mathbb{C}$ to get all matching pairs $\mathcal{M}_{\mathbb{C}} = \{ \mathcal{M}_{c} \}, c \in \mathbb{C}$.
	We use $\mathcal{M}_{\mathbb{C}}$ to meta-pretrain our MetaRetriever.
	
	Through the procedure introduced above, the I/O times of support-query pairing are 2.
	First, it reads all data to get the categories of each instance to run the support-query pairing algorithm.
	Second, given the pairing results, it takes the second I/O time to re-range all data to construct support-query pair.
	For the episodic training, every random-sampling will lead to a I/O time while our method can reduce I/O times significantly.
	
	\section{Details of Model Implementation}
	
	Since MetaRetriever works in a sequence-to-squence manner, we use the encoder-decoder-style architecture to achieve this procedure.
	We choose T5-base architecture~\citep{DBLP:journals/jmlr/RaffelSRLNMZLL20} as our model with 220M parameters.
	We pretrain our model on our constructed pretraining dataset.
	We use Adam~\cite{Kingma2015adam} as the optimizer with learning rate $1e-4$ with the linear scheduling with a warming up proportion 6\%.
	As previous work~\citep{uie2022lu}, we randomly sample 10 SpotName and AssoName to preserve the generalization ability of the model when calculating $\mathcal{L}_{\text{retrv}}$ and $\mathcal{L}_{\text{gen}}$.
	For $\mathcal{L}_{\text{LM}}$, we set the corruption rate as 15\% and the average corrupting span length as 3 as T5~\citep{2020t5}. 
	We truncate the length of the input sequence to 128 during pretraining.
	In the pretraining phase, we set the step of the inner loop as $1$ and the inner step size $\alpha$ and the outer step size $\beta$ are set as $1e-4$.
	The inner-loop update step $J$ is set as 1.
	We adopt the filtered version of our pretraining corpus (6.9M instances) to pretrain MetaRetriever.
	We pretrain our model for both $4$ epochs with batch size 512 on 8 NVIDIA A100 80G GPUs with 56 hours.
	For the downstream fine-tuning, we apply the same grid searching strategy for hyper-parameters as UIE~\citep{uie2022lu} to find the optimal results.
	In the inference phase, we use the greedy search strategy to decode the final predicted SEL sequence.

	\section{Details of Downstream Task Datasets}
	
	\label{sec:dataset_stat}
	
	\begin{table}[t]
		\centering
		\resizebox{0.49\textwidth}{!}{
			\begin{tabular}{c|ccc|ccc}
				\toprule
				& |Ent| & |Rel| & |Evt| & \#Train & \#Val & \#Test \\
				\midrule
				ACE04 & 7     & -     & -     & 6,202  & 745   & 812  \\
				ACE05-Ent & 7     & -     & -     & 7,299  & 971   & 1,060  \\
				CoNLL03 & 4     & -     & -     & 14,041  & 3,250  & 3,453  \\
				ACE05-Rel & 7     & 6     & -     & 10,051  & 2,420  & 2,050  \\
				CoNLL04 & 4     & 5     & -     & 922   & 231   & 288  \\
				SciERC & 6     & 7     & -     & 1,861  & 275   & 551  \\
				ACE05-Evt & -     & -     & 33    & 19,216  & 901   & 676  \\
				CASIE & 21     & -     & 5     & 11,189  & 1,778  & 3,208  \\
				14res & 2     & 3     & -     & 1,266  & 310   & 492  \\
				14lap & 2     & 3     & -     & 906   & 219   & 328  \\
				15res & 2     & 3     & -     & 605   & 148   & 322  \\
				16res & 2     & 3     & -     & 857   & 210   & 326  \\
				\bottomrule
			\end{tabular}%
		}
		\caption{
			Detailed datasets statistics of all downstream IE tasks.
			|*| indicates the number of categories, and \# is the number of sentences in the specific subset.
		}
		\label{tab:dataset_stat}
	\end{table}%
	
	We conduct evaluation experiments on 4 IE tasks, 12 datasets, and the detailed statistic of each dataset is shown in \tablename~\ref{tab:dataset_stat}.
	
	\section{Details of Effect of Finetune Epoch}
	
	\label{sec:details_effect_finetune_epoch}
	
	\begin{figure*}
		\begin{minipage}[t]{0.5\linewidth}
			\centering
			\includegraphics[scale=0.5]{finetune_epoch_conll03.pdf}
			\label{fig:appendix-fintune-epoch-conll03}
		\end{minipage}%
		\begin{minipage}[t]{0.5\linewidth}
			\centering
			\includegraphics[scale=0.5]{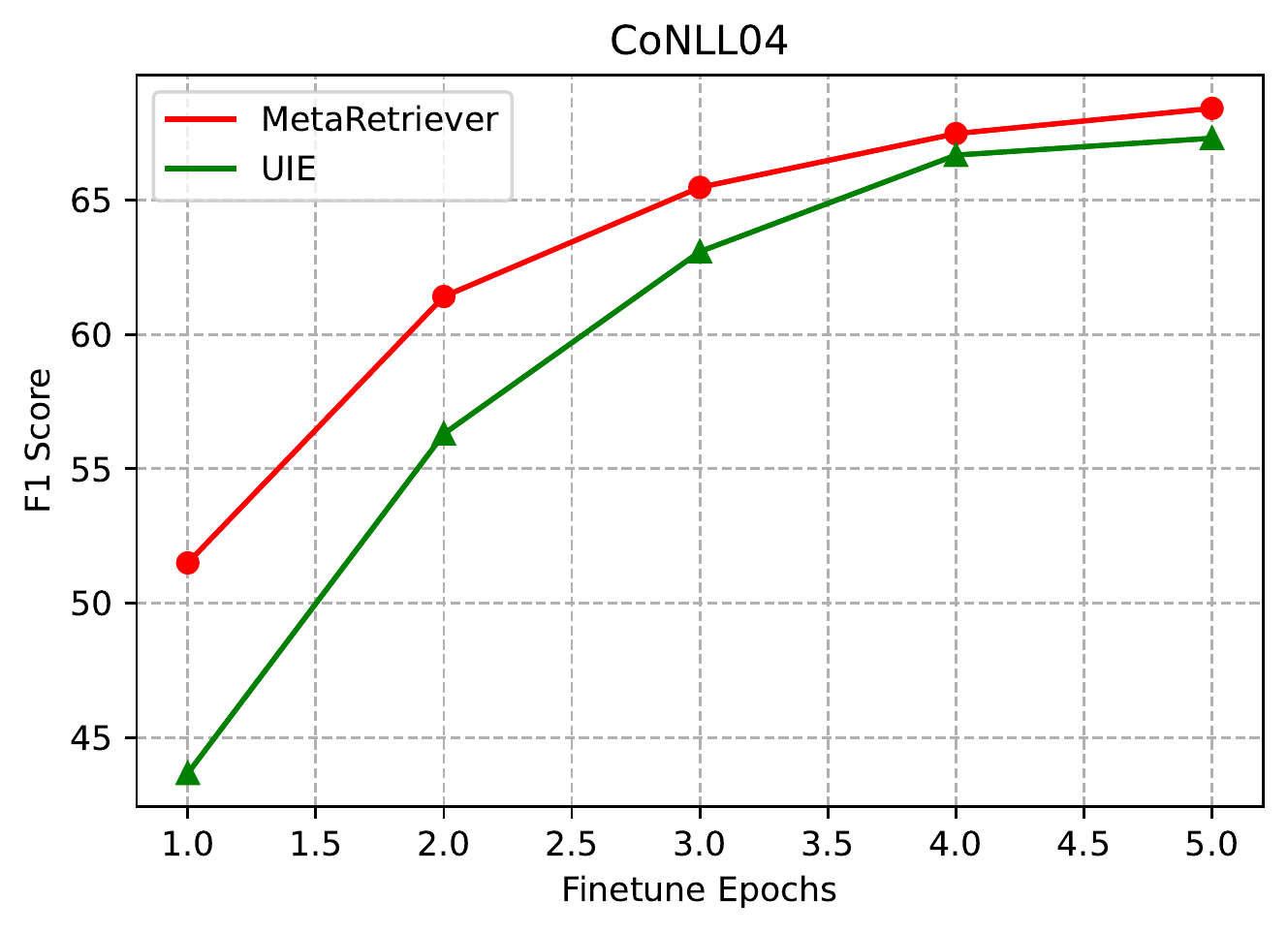}
			\label{fig:appendix-finetune-epoch-conll04}
		\end{minipage}
		\caption{Effect of Finetune Epoch.}
		\label{fig:appendix-finetune-epoch}
	\end{figure*}
	
	To verify that our MetaRetriever can fast learn the downstream IE tasks, we conduct experiments on both CoNLL03 and CoNLL04.
	Experimental results are shown in \figurename~\ref{fig:appendix-finetune-epoch}.
	Results of CoNLL04 show the same trend as CoNLL03, which can prove the effect of our proposed Meta-Pretraining Algorithm.
	
	\section{Details of Effect of Structure Complexity}
	
	\label{sec:details_effect_complicated_structure}
	
	\begin{figure*}
		\begin{minipage}[t]{0.5\linewidth}
			\centering
			\includegraphics[scale=0.5]{complicated_structure_conll03.pdf}
			\label{fig:appendix-complicated-structure-conll03}
		\end{minipage}%
		\begin{minipage}[t]{0.5\linewidth}
			\centering
			\includegraphics[scale=0.5]{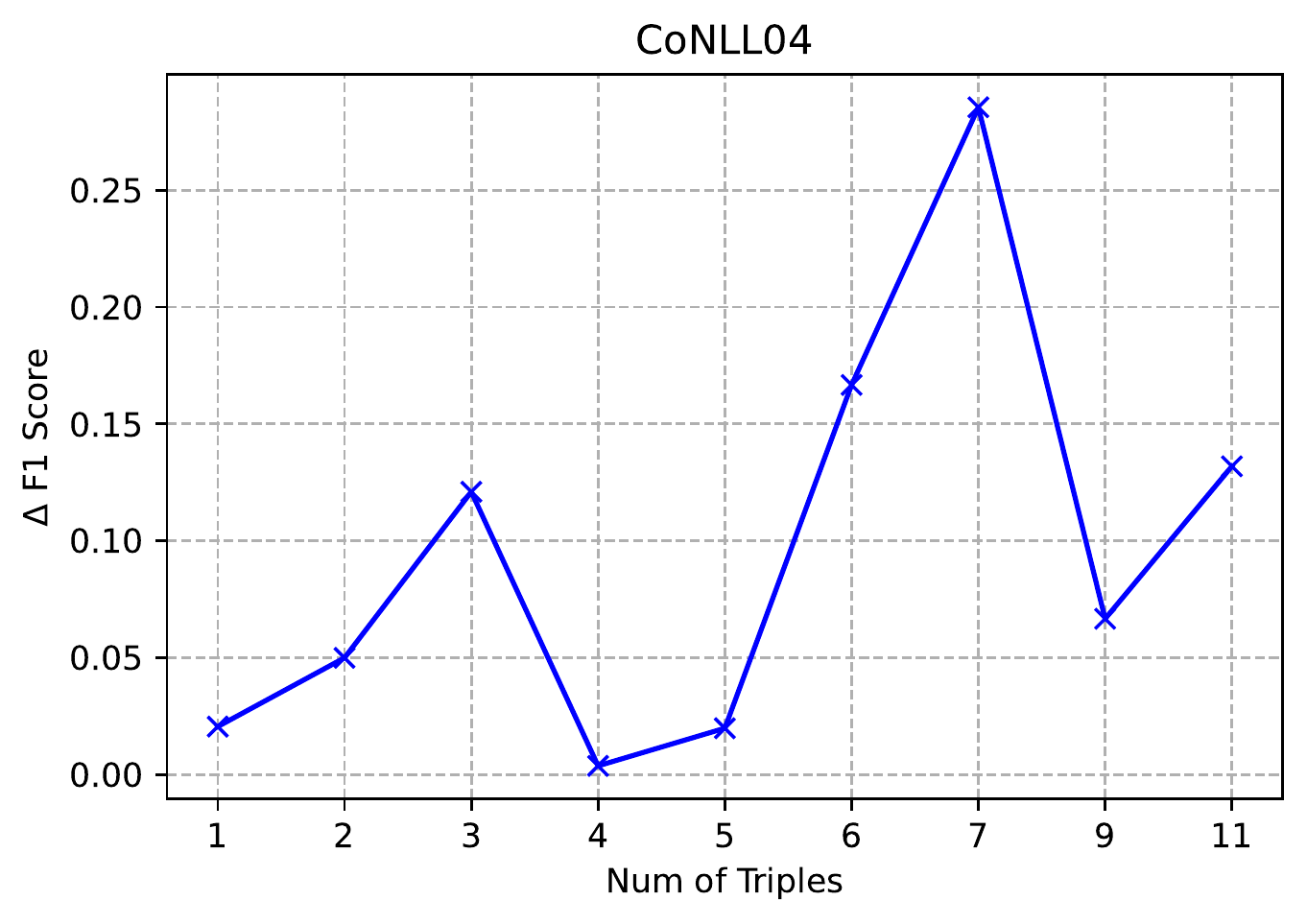}
			\label{fig:appendix-complicated-structure-conll04}
		\end{minipage}
		\caption{Effect of Structure Complexity.}
		\label{fig:appendix-complicated-structure}
	\end{figure*}
	
	To validate that MetaRetriever own the superiority when there exist complicated structures needed to be extracted.
	We group the test set of CoNLL03 and CoNLL04 according to the number of entities and the number of relational triples and calculate the performance difference between MetaRetriever and UIE in different group.
	Experimental results are shown in \figurename~\ref{fig:appendix-complicated-structure}.
	From it, we can learn that when there exist more entities and relational triples, MetaRetriever performs better.
	
	\section{Discussion of Efficiency}
	\label{sec:efficiency}
	
	While our MetaRetriever has achieved the new state-of-the-art on 4 IE tasks, 12 datasets, in fully-supervised, few-shot and low-resource scenarios, it still has several limitations on training efficiency and inference efficiency.
	
	\begin{table}[h]
		\centering
		\begin{tabular}{l|rr}
			\toprule
			& \multicolumn{1}{c}{UIE} & \multicolumn{1}{c}{MetaRetriever} \\
			\midrule
			CoNLL04 & 0.83s & 1.46s \\
			\bottomrule
		\end{tabular}%
		\caption{
			Inference time of MetaRetriever and UIE on the test set of CoNLL04 dataset.
		}
		\label{tab:efficiency_inference}
	\end{table}%
	
	First, since MetaRetriever will first retrieve task-specific knowledge and then make predictions in the inference phase, such a retrieve-then-extract manner will take longer inference time than non-retrieve methods unavoidably.
	%
	% It will cost about twice time as previous work to make predictions.
	%
	We conduct experiments on the test set of CoNLL04 dataset to compare overall inference time of MetaRetriever with UIE.
	All hyper-parameters are set to be the same for a fair comparison.
	Experimental results are shown in \tablename~\ref{tab:efficiency_inference} and we can find that MetaRetriever cost nearly twice time as UIE to make predictions. 
	As MetaRetriever works in a retrieve-than-extract manner, such a time cost is reasonable.

	\begin{table}[h]
		\centering
		\begin{tabular}{l|rr}
			\toprule
			& \multicolumn{1}{c}{SimpleRetriever} & \multicolumn{1}{c}{MetaRetriever} \\
			\midrule
			10K & 2min & 2min25s \\
			\bottomrule
		\end{tabular}%
		\caption{
			Pretraining time of MetaRetriever and SimpleRetriever on 10K instances for one epoch.
		}
		\label{tab:efficiency_pretraining}
	\end{table}%
	
	Second, our proposed meta-pretraining algorithm is based on bi-level optimization.
	In the pretraining phase, it needs to calculate high-order gradients to optimize parameters and calculating high-order gradient requires more time.
	Therefore, it takes longer time to pretrain MetaRetriever.
	To illustrate the time cost, we conduct experiments on 10K instances to compare the pretraining time of MetaRetriever with SimpleRetriever which is pretrained without meta-pretraining algorithm.
	All hyper-parameters are set to be the same for a fair comparison.
	\tablename~\ref{tab:efficiency_pretraining} gives the results.
	From it, we can obverse that compared with SimpleRetriever, MetaRetriever takes about 1/4 longer time than SimpleRetriever. 
	Finally, We spent 56 hours to pretrain MetaRetriever on filtered pretraining corpus (6.9M instances).
	
\end{document}